\documentclass{article}

\PassOptionsToPackage{numbers, compress}{natbib}

\usepackage[preprint]{neurips_2024}

\usepackage[utf8]{inputenc} 
\usepackage[T1]{fontenc}    
\usepackage{hyperref}       
\usepackage{url}            
\usepackage{booktabs}       
\usepackage{amsfonts}       
\usepackage{nicefrac}       
\usepackage{microtype}      
\usepackage{xcolor}         
\usepackage{graphicx}
\usepackage{amsmath}
\usepackage{subcaption}
\usepackage{booktabs}
\usepackage{multirow}
\usepackage{enumitem}
\usepackage{colortbl} 

\title{Knowledge Tagging System on Math Questions via LLMs with Flexible Demonstration Retriever}

\author{%
Hang Li$^{1,2}$ \quad Tianlong Xu$^{2}$ \quad Jiliang Tang$^2$ \quad Qingsong Wen$^{1}$\thanks{Corresponding author}\\
$^1$Squirrel AI \quad $^2$Michigan State University\\
\texttt{\{tianlongxu, qingsongwen\}@squirrelai.com}\\
\texttt{\{lihang4, tangjili\}@msu.edu}
}

\begin{document}

\maketitle

\newcommand{\jt}[1]{\textbf{\textcolor{red}{[JT: #1]}}}

\begin{abstract}

Knowledge tagging for questions plays a crucial role in contemporary intelligent educational applications, including learning progress diagnosis, practice question recommendations, and course content organization. 
Traditionally, these annotations are always conducted by pedagogical experts, as the task requires not only a strong semantic understanding of both question stems and knowledge definitions but also deep insights into connecting question-solving logic with corresponding knowledge concepts. 
With the recent emergence of advanced text encoding algorithms, such as pre-trained language models, many researchers have developed automatic knowledge tagging systems based on calculating the semantic similarity between the knowledge and question embeddings.
In this paper, we explore automating the task using Large Language Models (LLMs), in response to the inability of prior encoding-based methods to deal with the hard cases which involve strong domain knowledge and complicated concept definitions. 
By showing the strong performance of zero- and few-shot results over math questions knowledge tagging tasks, we demonstrate LLMs' great potential in conquering the challenges faced by prior methods. 
Furthermore, by proposing a reinforcement learning-based demonstration retriever, we successfully exploit the great potential of different-sized LLMs in achieving better performance results while keeping the in-context demonstration usage efficiency high. Data and code is available in \url{https://anonymous.4open.science/r/KnowTS-0563}.

\end{abstract}

\section{Introduction}

Knowledge concept tagging aims to generate a precise knowledge index to educational materials. It has been recognized as an important factor of current intelligent education systems in providing high-quality educational content to educators and learners during the practice~\citep{chen2014tag}. For example, with well-annotated education materials, teachers can enjoy great conveniences in organizing coursing content through searching concept keywords index~\citep{sun2018automatic}. Among the tagging objects, concept tagging over math questions has attracted greatly attention because of the recent successes of applying intelligent tutoring systems (ITS) in mathematical education~\citep{burns2013intelligent}. Traditionally, the questions' concept tags are annotated by the pedagogical experts. However, the rapid growth of the Internet has caused conventional manual methods to be insufficient to meet the demand for handling large volumes of online question data or updating existing concept tags in a timely fashion.

To solve the above issues, existing works~\citep{sun2018automatic,zhang2021question} have tried to automate the tagging process with different natural language processing (NLP) algorithms. 
For example, Du et al.\cite{du2021application} use text embedding techniques to convert the knowledge definitions and question stems into dense vectors, and then train machine learning classifiers based on the embedding similarities. However, such practice focuses only on comparing the explicit text semantic information but dismisses the implicit relationship in question solutions and knowledge concepts. It can cause unsatisfactory performance when faced with complicated knowledge descriptions and questions. One recent study attempts to improve the tagging performance by leveraging pre-trained language models (PLMs) and fusing external information, such as solution text and conceptual ontology, with original question contents during the judging process~\citep{huang2023pqsct}. Although this new trial demonstrates its effectiveness in solving the challenges faced by prior embedding-based methods, it introduces additional data requirements, e.g., complementary solution text to questions and conceptual ontology information between knowledge concepts, to the knowledge tagging model, which restrict the wide applications of the algorithm to questions with limited external resources.

In this work, we propose a novel knowledge-tagging framework KnowTS. It can leverage the advanced mathematical and logical inference capabilities~\citep{achiam2023gpt} of LLMs to enable knowledge tagging with only knowledge definition text. In addition, owing to the strong in-context learning ability of LLMs, KnowTS has the potential to be swiftly applied with a few annotation samples, setting it apart from all previous training-based algorithms. This feature allows KnowTS to be rapidly adapted for annotating works encompassing nearly all knowledge concepts and questions. Furthermore, due to the huge performance gap among using different sets of demonstration samples~\citep{wang2023large}, we propose a novel reinforcement learning (RL) based demonstration retriever focusing on dynamically providing flexible lengths of demonstration samples to every question knowledge matching queries. To validate the effectiveness of KnowTS, we experiment with an expert-annotated knowledge concept question dataset collected from a public K-12 education platform. Experimental results demonstrate that KnowTS can achieve the best in-context learning performance while using fewer demonstrations. 


\begin{figure}
    \centering
    \includegraphics[width=0.9\textwidth]{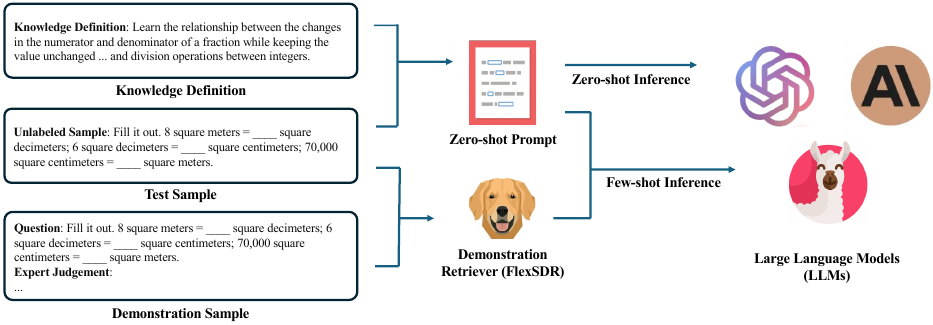}
    \vspace{-2mm}
    \caption{An overview of the workflow of the proposed KnowTS system.} 
    \label{fig:framework}
     \vspace{-5mm}
\end{figure}

\section{Related Work} \vspace{-2mm}


\begin{itemize}[leftmargin=*]
\item \textbf{Knowledge Concept Tagging}:
The major challenge of knowledge tagging tasks is how to construct a meaningful link in between the knowledge concepts and the problems, either through the description of the problem themselves or through solutions. The task formulation can primarily be categorized into two directions: retrieval and matrix decomposition. The former relies heavily on training a semantic representation. Sun et al. \cite{8295250} employs simple backbone models such as long short-term memory (LSTM) and some attention mechanisms to learn short-range dependency embeddings, where the questions are fed into LSTM layers and are ultimately connected to cross-entropy functions that indicate whether or not a tagging concept belongs to a given problem. Liu et al. \cite{liu2019ekt} devised an exercise-enhanced recurrent neural network with Markov property and Attention mechanism to extract rich knowledge concepts information in the exercise's content. Similarly but with enriched data source such as text,  multi-modal data~\citep{yin2019quesnet} as well as latex formula combined data \citep{huang2021context}, semantic representations learned with LSTM have been improved to capture more implicit contexts. Huang et al. \cite{huang2020neural} fills knowledge graph information into the embedding layers and achieves better mathematical semantic understanding. To take advantage of the robust transformers framework, Zemlyanskiy et al. \cite{zemlyanskiy2021docent} pretrained a BERT model to learn jointly predicting words and entities as movie tags given the reviews of movies. Huang et al. \cite{10123979} proposes an improved pretrained bidirectional encoder representation from transformers (BERT) for concept tagging with both questions and solutions. 


\item \textbf{In-context Learning Retriever}:
Few-shot in-context learning (ICL) is the ability of large language models (LLMs) to perform a new task when a few input-output examples or demonstrations for the new task are given alongside the actual task input~\citep{brown2020language}. Importantly, the model parameters do not have to be fine-tuned towards the new task. However, the performance of ICL varies significantly based on the choice of demonstrations~\citep{wang2023large}. In order to keep the stability of ICL performance and exploit the potential of LLMs, many ICL methods have been proposed in recent studies. Rubin et al. \cite{rubin2021learning} investigated using SBERT~\citep{reimers2019sentence} embeddings for demonstration retrieval and show that retrieving demonstrations based on SBERT embeddings often provides a boost in performance compared to zero-shot or random few-shot selection. Liu et al. \cite{liu2021makes} found that fine-tuning pre-trrained language models on task-related datasets offered further empirical gains. Besides using embedding to recall relevant demonstrations, recent works tried to build a proxy scoring model to score each candidate demonstration. Then, a retriever is trained, which separates top-score examples from bottom-score examples~\citep{rubin2021learning}. At last, Scarlatos and Lan \cite{scarlatos2023reticl} and Lu et al. \cite{lu2022dynamic} directly use response correctness as the reward function and train policy network with RL-method to decide the best demonstration for different samples dynamically.

\end{itemize}

\section{Method} \vspace{-2mm}

Before diving into the details of the method, we first give a formal problem definition to the knowledge tagging task as follows: Given a pair of knowledge definition text $k$ and a question's stem text $q$, the objective of a concept tagging model is to produce a binary judgment $y\in\{0,1\}$, where 1 means $k$ and $q$ are matching, 0 otherwise. In the following subsections, we first present an overview of our proposed framework, KnowTS. Then, we introduce details about the implementations of KnowTS with zero-shot and few-shot inference pipelines. Lastly, to further boost the performance of KnowTS with demonstration samples, we propose FlexSDR, an RL-based retriever algorithm, to achieve efficient and high-performance knowledge tagging.





\subsection{An Overview}
\vspace{-2mm}

An overview of KnowTS is demonstrated in Fig.~\ref{fig:framework}. It consists of three key components: (1) zero-shot inference pipeline, (2) few-shot inference pipeline, and (3) adaptive demonstration retriever. Each component in KnowTS plays different roles to accommodate different knowledge tagging scenarios. When there is no available annotated data, KnowTS will leverage a zero-shot pipeline to generate the judgment directly. And, when there are limited available demonstration examples, KnowTS will use the few-shot inference via its in-context learning capability. When the demonstration samples are needed to select for demonstration, KnowTS will utilize a demonstration retriever to adaptively select effective demonstrations for different $(k,q)$ pairs.

\subsection{Zero-shot and Few-shot Pipelines of KnowTS}
\label{sec:zeroshot}
\vspace{-2mm}


One key difference between KnowTS and other prior machine learning models is its strong performance while facing limited or even no annotated data for each knowledge $k$. Such advantages contribute to the powerful zero-shot inference capability of LLMs, which is empowered by its huge size model parameter and the extensive pre-training on diverse and vast datasets. Prior studies by Wei et al. \cite{wei2021finetuned} demonstrate that LLMs have strengths in comprehending instructions in natural language and applying learned knowledge to new problems with limited or even no training data specific to these tasks. In our problem, we leverage this capability by composing a zero-shot prompt as follows: we first describe the goal of the tagging task as \emph{"You are a knowledge concept annotator. Your job is to judge whether the <Question> is concerning the <Knowledge>."} Then, for the convenience of the processing procedure, we add a response format instruction in the prompt: \emph{"The judgment token: <Yes> or <No> should be provided at the end of the response."} At last, as the prior studies like Chain-of-Thought (COT)~\citep{wei2022chain} have discovered, instructing LLMs to generate step-by-step problem-solving solutions will be helpful for the LLMs to draw the correct conclusions, especially when faced with complicated problems, we ask LLMs to not only provide their positive or negative predictions but also present the reason at first: \emph{"You should first provide the reasons before giving your judgement."} Overall, the zero-shot task instruction prompt is presented as follows:
\begin{quote}

\emph{\textbf{Instruction}}: You are a knowledge concept annotator. 
Your job is to judge whether the <Question> is concerning the <Knowledge>. 
You should first provide the reasons before giving your judgment. 
The judgment token: <Yes> or <No> should be provided at the end of the response. 

\emph{\textbf{Knowledge}:} \underline{<Knowledge>: The composition of numbers within 20.}

\emph{\textbf{Question}:} \underline{<Question>: There are (\ ) tens and (\ ) ones in 14.}

\emph{\textbf{Judgement}:} (Generated by LLMs)

\end{quote}
Although the zero-shot prompt provides a promising solution without using any annotated samples,  the knowledge definition text $k$ may sometimes not be specific enough for complicated judgments. For example, there is a knowledge concept named \emph{consecutive carry in multiplication}, which occurs when the product of two digits, along with any carry from the previous calculation, results in a number greater than 9, thus requiring another carry to be added to the next column in the calculation. It is hard for many LLMs to catch that key point during the judging process without any hints. To overcome this problem, KnowTS can leverage a few-shot inference pipeline when there are available demonstration samples associated with the given knowledge $k$. Contributing to LLMs' strong in-context learning (ICL) capability, KnowTS can imitate the judging logic of the given demonstrations and achieve significant performance gain even with limited samples provided. Detailed comparisons between zero-shot and few-shot responses are shown in Section~\ref{sec:fewshot}. Below, we give a demonstration example, which interprets the "consecutive carry" to LLMs with an example judgment written by experts $38 \times 9$:


\begin{quote}
    There is a consecutive carry start from multiplication of the one's place ($8 \times 9 = 72$, carry $7$), and then a carry from the tens place operation ($3 \times 9 + 7 = 34$, carry $3$). Thus, the question matches the given knowledge descriptions. <Yes> 
\end{quote}

Finally, by presenting the example answers as inputs and instructing LLMs to generate similar responses, LLMs will learn and follow the judging steps of demonstrations, which will also help them output more relevant responses. 


\subsection{Flexible Sequential Demonstration Retriever (FlexSDR)}

Although incorporating available demonstrations has the potential to bring LLMs performance gain compared to the zero-shot setting, the effectiveness of each demonstration varies~\citep{liu2021makes}. Moreover, different input pairs $(k,q)$ may prefer to different combinations of demonstrations. To fully exploit the potential of KnowTS, we propose a reinforcement learning (RL) based demonstration selection method, termed Flexible Sequential Demonstration Retriever (FlexSDR), aiming to help LLMs exploit their potential from the demonstration samples while keeping only the necessary demonstrations as input for each input query. Formally, we define the key components of Markov Decision Process (MDP) in our problem as follows: Given the $t$ step's status $s_t = \{k, q, e_1, .., e_{t-1} \}$, where $(k,q)$ are the input knowledge and question pair and $e_{i|1 \leq i \leq t-1}$ are the demonstrations selected in prior $t-1$ steps, we hope to use the policy $\pi$ to generate the subsequent action $a_t$, which selects one demonstration sample $e_{t} \in E_{\mathcal{D}}$, where $E_{\mathcal{D}}$ is the demonstration bank or choose the stop signal $e_{\mathcal{E}}$. This process's reward is finding the best demonstration sequence (action trajectory $\tau$) which helps LLMs correctly judge the knowledge matching $(k,q)$ while keeping the $|\tau|$ small. To be noticed although there are several  RL-based methods, we point out that FlexSDR has its novelty in two perspectives: (1) we introduce the "early stop" option to each interactive step and use the stop bonus reward to guide the policy network $\pi$ to learn when to stop during the reinforcing process. With such design, FlexSDR avoids retrieving redundant demonstrations because of the prefixed demonstration size parameter and reduces the risk of sub-optimal few-shot inference performance~\citep{zhao2023dynamic}; and (2) We incorporate each intermediate step as an individual training sample and help the policy network learn to conduct the best action decision based on each step's response correctness status. More details about the effectiveness of each design are discussed in Sec~\ref{sec:abla}. In the following subsections, we introduce both the policy network and reward design of FlexSDR in details.

\begin{figure}
    \centering
    \includegraphics[width=0.99\textwidth]{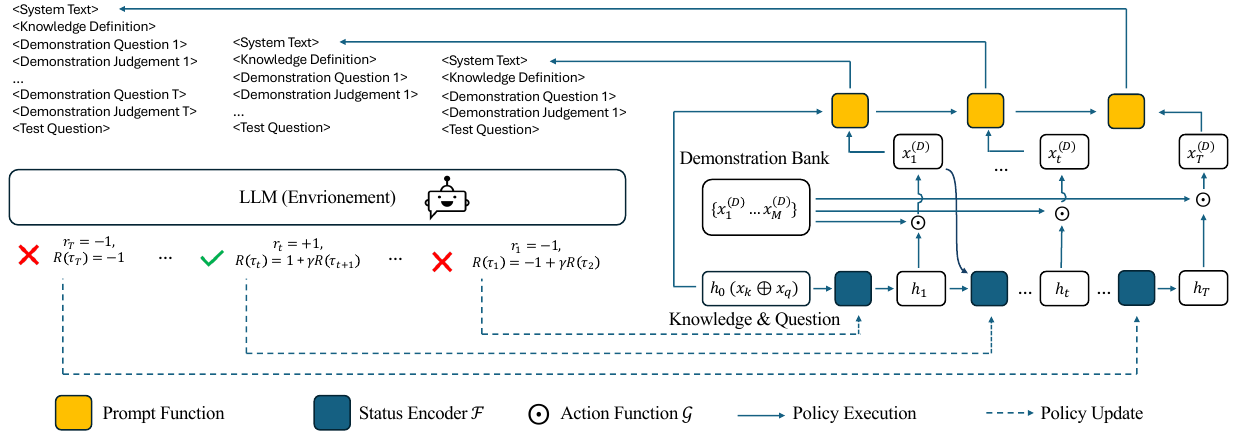} \vspace{-2mm}
    \caption{The framework of the proposed FlexSDR.}
    \label{fig:flexsdr}
    \vspace{-5mm}
\end{figure}

\subsubsection{Policy Network}

Firstly, we define the policy network as $\pi_\theta$, and the policy execution process is to select an action $a_t$ from the probability distribution calculated by $\pi_\theta(a|s_t)$. As the action space for our policy network is decided by the demonstration bank $E_{\mathcal{D}}$, we decompose $\pi_\theta$ into two components $\pi_\theta = \mathcal{G}(\mathcal{F}(s_t), [E_{\mathcal{D}}\|e_\mathcal{E}])$, where $\mathcal{F}$ is a status encoder function that converts the sequential based status variable $s_t$ into a status vector $h_t$. $\mathcal{G}$ is an action function that calculates each action score for each available demonstration sample. $[\cdot\|\cdot]$ is the concatenation operation, and $e_\mathcal{E}$ is the early stop option. Following the prior work's setting~\citep{scarlatos2023reticl}, we choose to use the long short-term memory (LSTM) model~\citep{graves2012long} as $\mathcal{F}$ and a bilinear transformation as $\mathcal{G}$. Overall, the policy execution process is shown as the right part of Fig.~\ref{fig:flexsdr}, where the input $(k, q)$ pair is first encoded and used as the initial inputs for $\mathcal{F}$. Then, the $t$th-step hidden state output $h_t$ is used by $\mathcal{G}$ to calculate the selection score for each available demonstration and the early stop option.  After that, the action $a_t$ will be selected based on the scores. If the demonstration is selected, it will be appended to the prompt and interact with LLMs to calculate the $t$-th step reward score. The process ends whenever the early stop option is hit or reaches the max-allowed length. Formally, the procedure can be defined as follows: 
\begin{align}
   & x_e = \mathcal{E}(e),\ \ x_k = \mathcal{E}(k),\ \ x_q = \mathcal{E}(q), X_{E_{\mathcal{D}}} = \mathcal{E}(E_{\mathcal{D}})\\
   & h_t = \mathcal{F}(s_t) = \mathrm{LSTM}(h_0;x_{e_1},...,x_{e_{t-1}}),\ \ h_0 = \mathrm{tanh}(W_0[x_k\|x_q] + b_0)\\
   & \pi_\theta(a|s_t) = \mathrm{Softmax}(\mathcal{G}(\cdot)),\ \ \mathcal{G}(h_t, [X_{E_{\mathcal{D}}}\|x_{x_{\mathcal{E}}}] ) = h_t W_a [X_{E_{\mathcal{D}}}\|x_{x_{\mathcal{E}}}]^T
\end{align}
where $x_e, x_k, x_q$ are the encoding results of knowledge text, question text, and demonstration text. $\mathcal{E}$ is the pre-trained text encoding model. $W_0$ and $b_0$ are the parameters of knowledge and question information fusing layer, and $W_a$ is the parameter of bilinear transformation.

\subsubsection{Learning Rewards}

\begin{figure}[!t]
\centering
\begin{subfigure}{.5\textwidth}
  \centering
  \includegraphics[width=.99\linewidth]{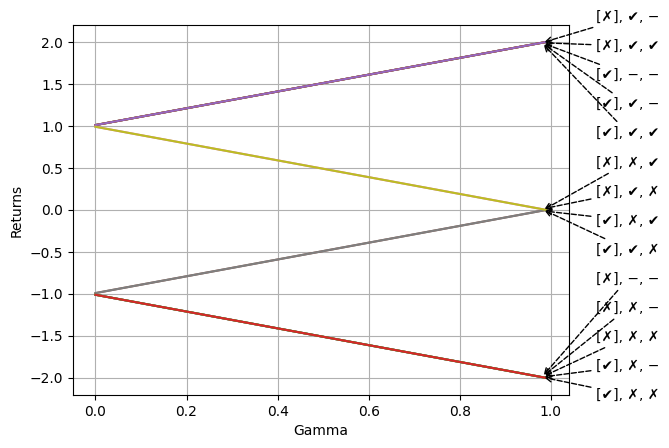}
  \caption{$R(\tau)$ with $\gamma \in (0,1)$}
  \label{fig:return_nob}
\end{subfigure}%
\begin{subfigure}{.5\textwidth}
  \centering
  \includegraphics[width=.99\linewidth]{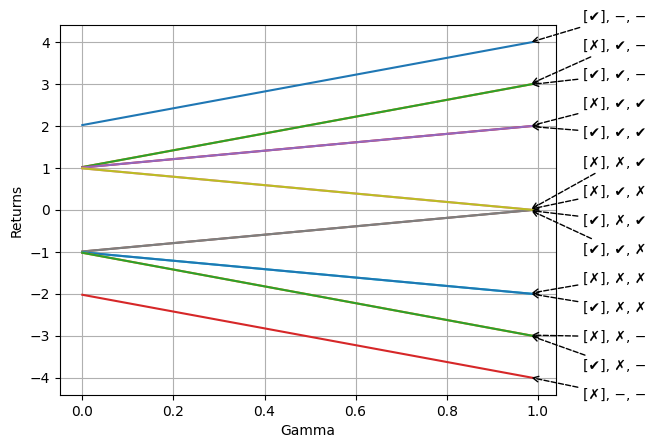}
  \caption{$R'(\tau)$ with $\omega=1,\ \gamma \in (0,1)$}
  \label{fig:return_b}
\end{subfigure}
 \vspace{-4mm}
\caption{Return functions w/o and w/ stop bonus reward where $T=2$.}
\label{fig:return} \vspace{-5mm}
\end{figure}

To train the policy network $\pi_\theta$, we use the proximal policy optimization (PPO) method ~\citep{sutton2018reinforcement}. To be specific, we define the step-wised reward function of FlexSDR as follows:
\begin{equation}
    r_t =  \mathrm{EVAL}(\hat{y}_t, y),\ \ r_t \in \{-1,1\}
\end{equation}
where $\mathrm{EVAL}$ is the evaluation function that compares the judgment response by LLMs $\hat{y} = \mathrm{LLM}(k,q,e_1,...,e_t)$ with the expertise judgment $y$. If the two judgments are the same, the reward for timestep $t$ will be $+1$. Otherwise, the reward value will be $-1$. For early-stop actions $e_\mathcal{E}$, we calculate its correctness based on its most recent step. This reward design differs from previous RL-based retriever training algorithms~\citep{scarlatos2023reticl}, which calculate the reward only at the final timestep $T$. At this point, the size of the retrieved demonstration reaches its maximum allowable limit. For FlexSDR, we calculate rewards $r_t$ for all the timesteps and use the discounted trajectory return $R(s_t,a_t) = r_t + \gamma R(s_{t+1}, a_{t+1})$, where $\gamma \in (0,1)$ is the discount factor, to calculate the action returns along the trajectory $\tau$, presented in left part of Fig.~\ref{fig:flexsdr}. The final goal of our optimization is to maximize the expectation return of the trajectory $\tau$ generated by the iterative execution on the optimized policy network $\pi_\theta$: 
\begin{equation}
    J_\theta = E_{\tau \sim p_{\pi_{\theta}}(\tau)}[\Sigma_t R(s_t, a_t)]
\end{equation}
As the instant reward $r_t$ can only be two values -1 or +1, and the maximum length of allowed demonstration for RL-based retriever training is limited, we enumerate all the possible cases for different types of correctness status across the fixed length trajectory ($T=2$) in Fig~\ref{fig:return_nob}. By viewing the corresponding rewards for different trajectories, we can have the following observation: (1) due to the existing discount factor $\gamma$, the trajectory with earlier steps approaching the correct response will tend to have a higher reward: $([\times],\checkmark,\checkmark) > ([\times],\times,\checkmark)$, which encourages the policy network to find the most valid demonstration at each iteration; (2) when the policy makes an error attempt at the future steps, its return $R(\tau)$ will be decreased, e.g., $([\times],\checkmark,\checkmark) > ([\times],\checkmark,\times)$, this instructs the policy network to avoid appending the inappropriate demonstrations. In addition to the correct reward, we introduce another "stop bonus" reward $r'_t$ to each time step: 
\begin{equation}
    r'_t =
    \begin{cases}
        0,\ \ \ \ \ \ a_t \ne e_{\mathcal{E}}\\
        r_{t-1},\ a_t = e_{\mathcal{E}}
    \end{cases}
\end{equation}
The bonus added trajectory return can be written as $R'(s_t, a_t) = (r_t + \omega * r'_t) + \gamma R'(s_{t+1},a_{t+1})$, where $\omega$ is a weight parameter balancing the influence of the stop bonus to the final returns. The stop bonus trajectory reward function is plotted in Fig~\ref{fig:return_b}. From the plot, we can observe that $R'(\tau)$ with the earlier correct stop action $(T,S)$ will receive a higher return compared to the ones with keeping selecting the demonstrations: $([\checkmark],-,-) > ([\checkmark],\checkmark,-) > ([\checkmark],\checkmark,\checkmark)$. This reward design will encourage the policy network to stop early when the correct response is met. For the case when the stop action is given after an error attempt, the return will be penalized: $([\times], -,-) < ([\times],\times,-) < ([\times],\times,\times)$, since it stops exploring the other possibility for finding the correct response.

\section{Experiment}
In this section, we conduct experiments to validate the effectiveness of each component in KnowTS. Through the experiments, we aim to answer the following research questions:
\vspace{-2mm}
\begin{itemize}[leftmargin=*]
    \item RQ1: Can zero-shot and few-shot pipelines of KnowTS help it outperform prior machine learning models while facing limited or even no annotated data?
    \item RQ2: Can FlexSDR further boost the few-shot pipeline performance while using fewer demonstrations compared to the other RL-based Retrievers?
    \item RQ3: Which components of FlexSDR are effective?
    \item RQ4: What's the evidence supporting FlexSDR learns the correct stop time from the training?
\end{itemize}
\subsection{Dataset Overview}

To answer the research questions above, we collect a knowledge concept tagging dataset, MathKnowCT, from an online K-12 math education platform. The dataset consists of 24 knowledge concepts, spreading from the math concept learning goal of Grade 1 to Grade 3 students. For each knowledge concept, we collect 100 candidate questions from an unlabeled question database with the highest text embedding similarity and then ask a pedagogical expert to conduct the matching annotations. The ratio between matching and mismatching categories of the whole dataset is around 1:4. More details about the dataset statistics and knowledge concept definitions can be found from Table~\ref{tab:data_detail} in Appendix~\ref{sec:data_detail}. Before the experiment, we first split 5 positive samples and 5 negative samples for each knowledge concept as the training (demonstration) set. For each sample in the training (demonstration) set, we ask a pedagogical expert to complete the reasons for the judgment. 

\subsection{Baselines}


To answer RQ1, we prepare baseline models as follows:
\vspace{-2mm}
\begin{itemize}[leftmargin=*]
    \item \textbf{Embedding Similarity}: we first use two high-performed long text encoding models, sentence-BERT(S-BERT) ~\citep{reimers2019sentence} and text-embedding-3-small \footnote{\url{https://platform.openai.com/docs/guides/embeddings/embedding-models}}, to encode both $k$ and $q$ into dense embedding vectors $x_k$ and $x_q$ and we calculate the cosine similarity between them. The judgment of each test sample is determined by the top-$K$ selection or similarity threshold $\eta$ comparisons. The value of hyper-parameter $K$ and $\eta$ is determined by performing a grid search on test data.
    \item \textbf{PLM Fine-tuning}: We add a binary classification layer to the top of <BOS> tokens outputs and fine-tune the parameter of the whole model with the binary entropy loss calculated on the samples in the training set. The PLMs we use in our experiment include BERT~\citep{devlin2018bert}, T5~\citep{raffel2020exploring}, and RoBERTa~\citep{liu2019roberta} and the learning rate during the fine-tuning process is tuned from 1e-3 to 1e-5.
\end{itemize}

To answer RQ2, we compare it with two prior SOTA RL-retrievers.
\vspace{-2mm}
\begin{itemize}[leftmargin=*]
    \item \textbf{PromptPG}~\citep{lu2022dynamic}: We implement PromptPG from the public available source code \footnote{\url{https://github.com/lupantech/PromptPG}}, where the retrieval is set as a one-layer-MLP . During the retrieval process, we input the input question and demonstration question embedding into the retriever and optimize the policy network with the REINFORCE policy gradient algorithm~\citep{williams1992simple} suggested by the paper.
    \item \textbf{RetICL}~\citep{scarlatos2023reticl}: RetICL can be viewed as a special case of FlexSDR, where no early stop is added to the demonstration selection space, decay factor $\gamma$ is set to 1 during the training, and the reward loss will only be calculated for the last step $T$'s result, we implement the algorithm based on changing the setting parameter of FlexSDR. The other training settings, e.g., retriever parameter sizes, are kept the same as FlexSDR for fair comparisons.
    
\end{itemize}

\subsection{LLMs Settings}

To validate the generosity of our proposed algorithm, we experiment with 4 representative LLMs frameworks for both the zero-shot and naive few-shot inference experiment, including GPT ~\citep{brown2020language}, LLAMA3~\citep{touvron2023llama}, Mistral and Mixtral~\citep{jiang2024mixtral}, Qwen1.5~\citep{bai2023qwen}. More details about LLM's implementation can be found in Appendix~\ref{sec:llm_implement}.


\subsection{FlexSDR Settings}

To implement FlexSDR, we choose to use a 2-layer LSTM with 64 hidden neurons for each layer, and the text encoder $\mathcal{E}$ is text-embedding-3-small, the early-stop bonus weight $\omega=1$. The discount factor $\gamma=0.99$. To improve the convergence of the whole training process, we employ the actor-critic optimization framework~\citep{konda1999actor} during training. We train the value function estimator using mean squared error (MSE) based on each step's hidden state $\mathcal{V}(h_t)$, the weight for the loss of value function is set as 0.5. Besides, to further improve the data usage efficiency, we also incorporated off-policy learning epochs during the training, and the off-policy epochs we set in our experiment is 80. Finally, to encourage exploration during the reinforcement steps, we add the negative entropy of the policy to each time step's loss, and the weight is set as 0.01. During the inference time, we use the greedy decoding method at each timestep $t$, and once the early stop option is hit, the demonstration retrieval procedure stops. 



\subsection{Zero-Shot and Naive Few-Shot Results}
\label{sec:zero-shot}

\begin{table}[]
\caption{Comparison between PLM Embedding Similarity, PLM Fine-tune, and LLM Zero-shot Inference. The best performance for each metric is marked with \textbf{bold}, and the second best one is marked with \underline{underline}.}
\label{tab:zero-shot}
\resizebox{\textwidth}{!}{
\begin{tabular}{@{}ccccccccccccc@{}}
\toprule
\multirow{2}{*}{Metric} & \multirow{2}{*}{\begin{tabular}[c]{@{}c@{}}Model \\ Size\end{tabular}} & \multicolumn{2}{c}{\begin{tabular}[c]{@{}c@{}}K / Q \\ Simularity\end{tabular}} & \multicolumn{2}{c}{\begin{tabular}[c]{@{}c@{}}Q / Q \\ Simularity\end{tabular}} & \multicolumn{3}{c}{\begin{tabular}[c]{@{}c@{}}PLM \\ Fine-tune\end{tabular}} & \multicolumn{4}{c}{\begin{tabular}[c]{@{}c@{}}LLM \\ Zero-Shot\end{tabular}} \\ \cmidrule(l){3-13} 
 &  & GPT-Embed & SBERT & GPT-Embed & SBERT & BERT & RoBERTa & T5 & GPT & Llama-3 & Mixtral & Qwen \\ \midrule
 \multirow{2}{*}{Accuracy} & Base & 67.43 & 79.9 & 78.52 & 63.58 & 58.45 & 35.51 & 77.18 & 75.30 & 58.44 & 68.60 & 66.89 \\
 & Large & - & - & - & - & 76.64 & 79.08 & 79.55 & \textbf{89.00} & 68.14 & 74.73 & \underline{79.98} \\ \midrule
\multirow{2}{*}{Precision} & Base & 52.68 & 67.66 & 67.51 & 49.1 & 44.03 & 35.51 & 64.70 & 60.09 & 46.18 & 53.87 & 52.58 \\
 & Large & - & - & - & - & 63.02 & \underline{72.61} & 71.45 & \textbf{78.38} & 52.58 & 59.89 & 65.34 \\ \midrule
\multirow{2}{*}{Recall} & Base & 75.27 & 82.39 & 75.4 & 87.63 & 62.77 & \textbf{100.00} & 78.63 & 89.25 & 92.37 & 89.74 & 83.16 \\
 & Large & - & - & - & - & 82.80 & 65.94 & 70.62 & 95.03 & \underline{98.79} & 85.89 & 94.74 \\ \midrule
\multirow{2}{*}{F1} & Base & 61.98 & 74.3 & 71.24 & 62.93 & 51.75 & 52.41 & 70.99 & 71.82 & 61.58 & 67.32 & 64.42 \\
 & Large & - & - & - & - & 71.57 & 69.12 & 71.03 & \textbf{85.91} & 68.63 & 70.57 & \underline{77.34} \\ \bottomrule
\end{tabular}}
\vspace{-5mm}
\end{table}

\begin{table}[]
\centering
\caption{Comparisons between LLM 2-Shot and 4-Shot Inference. The best performance for each metric is marked with \textbf{bold}, and the second best one is marked with \underline{underline}.}
\label{tab:few-shot}
\resizebox{\textwidth}{!}{
\begin{tabular}{@{}cccccccccccccc@{}}
\toprule
\multirow{3}{*}{Metric} & \multirow{3}{*}{Model Size} & \multicolumn{6}{c}{2-Shot} & \multicolumn{6}{c}{4-Shot} \\ \cmidrule(l){3-14} 
 &  & \multicolumn{3}{c}{Random} & \multicolumn{3}{c}{Heuristic} & \multicolumn{3}{c}{Random} & \multicolumn{3}{c}{Heuristic} \\ \cmidrule(l){3-14} 
 &  & GPT & Llama-3 & Mixtral & GPT & Llama-3 & Mixtral & GPT & Llama-3 & Mixtral & GPT & Llama-3 & Mixtral \\ \midrule
 \multirow{2}{*}{Accuracy} & Base & 76.01 & 75.64 & 78.72 & 72.50 & 73.15 & 81.15 & 77.95 & 79.25 & 81.33 & 79.56 & 80.74 & 80.62 \\
 & Large & 89.45 & 83.45 & 80.84 & 90.10 & 84.26 & 80.23 & \underline{90.40} & 88.00 & 84.31 & \textbf{91.11} & 88.57 & 84.06 \\ \midrule
\multirow{2}{*}{Precision} & Base & 60.33 & 60.59 & 65.11 & 57.22 & 58.47 & 68.86 & 62.65 & 64.91 & 68.44 & 64.98 & 67.52 & 67.94 \\
 & Large & 79.86 & 69.16 & 67.67 & 81.86 & 70.98 & 67.64 & \underline{82.56} & 76.03 & 73.18 & \textbf{83.86} & 77.91 & 73.35 \\ \midrule
\multirow{2}{*}{Recall} & Base & 93.41 & 89.82 & 86.31 & 87.37 & 84.14 & 85.64 & 92.88 & 90.48 & 87.98 & 91.26 & 88.15 & 85.98 \\
 & Large & 93.99 & \underline{95.83} & 87.50 & 92.65 & 93.68 & 84.27 & 92.49 & \textbf{96.37} & 87.63 & 92.82 & 94.35 & 87.63 \\ \midrule
\multirow{2}{*}{F1} & Base & 73.31 & 72.36 & 74.23 & 69.15 & 68.99 & 76.34 & 74.82 & 75.59 & 76.99 & 75.91 & 76.47 & 75.90 \\
 & Large & 86.35 & 80.34 & 76.32 & 86.92 & 80.76 & 75.04 & \underline{87.24} & 85.00 & 79.76 & \textbf{88.11} & 85.35 & 79.86 \\ \bottomrule
\end{tabular}}
\vspace{-5mm}
\end{table}

In this section, we answer RQ1 with comparisons between prior machine learning algorithms, including embedding similarity, PLM fine-tuning, and LLM-based methods, e.g., zero-shot, 2-shot, 4-shot inference. From Table~\ref{tab:zero-shot}, we can observe that the prior methods present acceptable performance. However, the performance of most of these methods got trapped at around 71\% F1-score. For LLM-based methods, we can find even under the zero-shot setting, some of the large-sized ones, e.g., GPT-4-turbo, present extremely strong task-solving capability and achieving 85.9\% F1 results, outperforming the non-LLM methods by a great margin. This observation proves our hypothesis that contributes to the broad prior knowledge (math concepts) learned during the pre-training phase and strong problem-solving skills taught in the instruction tunning stages, LLMs are good tools for knowledge tagging tasks with limited or even no annotation data. The result presented in Table~\ref{tab:few-shot} demonstrates the advantages of LLMs in-context learning capability. With the introduction of only 2 to 4 demonstration samples, most LLMs can achieve significantly better performance compared to the zero-shot cases, and LLMs with lower performance in zero-shot, e.g., Llama-3-70B, receive the performance boost by 10\%. Such observation suggests the great potential of LLM-based algorithms in generating high-performed knowledge tagging results with sufficient demonstration samples.  


\subsection{Demonstration Retriever Enhanced Few-shot Results}

\label{sec:fewshot}

\begin{table}[]
\centering
\caption{Comparisons between three RL-based retrievers on three LLMs. The best performance for each metric is marked with \textbf{bold}, and the second best one is marked with \underline{underline}. The number in (parentheses) for FlexSDR is the mean demonstration size the retriever decides.}
\label{tab:rl-shot}
\resizebox{\textwidth}{!}{
\begin{tabular}{@{}ccccccccccc@{}}
\toprule
\multirow{2}{*}{Metric} & \multirow{2}{*}{\begin{tabular}[c]{@{}c@{}}Max-Shot \\ Size\end{tabular}} & \multicolumn{3}{c}{GPT Base (GPT-3.5-turbo)} & \multicolumn{3}{c}{Llama-3 Base (Llama-3-8B)} & \multicolumn{3}{c}{Mixtral Base (Mistral-7B)} \\ \cmidrule(l){3-11} 
 &  & PromptPG & RetICL & FlexSDR & PromptPG & RetICL & FlexSDR & PromptPG & RetICL & FlexSDR \\ \midrule
\multirow{2}{*}{Accuracy} & 2 & 81.69 & 79.32 & \textbf{83.87} (1.38) & 70.68 & 79.89 & 79.32 (1.55) & 72.30 & 80.08 & \underline{83.68} (1.78) \\
 & 4 & 77.23 & 79.70 & 82.07 (2.10) & 73.15 & 77.89 & 80.65 (3.65) & 74.10 & 79.98 & 81.31 (2.82) \\ \midrule
\multirow{2}{*}{Precision} & 2 & 71.69 & 65.46 & \underline{73.54} (1.38) & 55.77 & 66.87 & 66.01 (1.55) & 57.89 & 69.77 & \textbf{77.81} (1.78) \\
 & 4 & 62.46 & 65.31 & 68.99 (2.10) & 57.96 & 63.15 & 66.67 (3.65) & 59.30 & 65.34 & 69.68 (2.82) \\ \midrule
\multirow{2}{*}{Recall} & 2 & 81.32 & 90.26 & 86.32 (1.38) & 90.26 & 87.63 & 87.89 (1.55) & 85.00 & 78.95 & 76.58 (1.78) \\
 & 4 & 92.37 & \underline{93.16} & 91.32 (2.10) & 92.89 & 92.89 & 92.63 (3.65) & 89.74 & \textbf{94.74} & 85.26 (2.82) \\ \midrule
\multirow{2}{*}{F1} & 2 & 76.20 & 75.88 & \textbf{79.42} (1.38) & 68.94 & 75.85 & 75.40 (1.55) & 68.87 & 74.07 & 77.19 (1.78) \\
 & 4 & 74.52 & 76.79 & \underline{78.60} (2.10) & 71.39 & 75.19 & 77.53 (3.65) & 71.41 & 77.34 & 76.69 (2.82) \\ \bottomrule
\end{tabular}}
\vspace{-3mm}
\end{table}

In this section, we answer RQ2 by presenting the comparisons between FlexSDR and other baselines in Table~\ref{tab:rl-shot}. From the table, we observe that FlexSDR constantly bring further boosts to the few-shot learning performance compared to the naive few-shot learning results in Table~\ref{tab:few-shot}. However, the performance gain of the other two baselines is not stable, especially PromptPG, which sometimes performs much worse than naive selected few-shot demonstrations. We believe the discrepancy is mostly due to the simple optimization goal of PromptPG, which treats each demonstration as an independent item and oversimplifies the retrieving process in a one-shot manner. Apart from that, as FlexSDR is designed with the early stop mechanism, the average demonstration length used in few-shot learning FlexSDR is always less than the max-shot size. From the table, we find FlexSDR uses 25\% less demonstrations for its few-shot learning inference, which achieves our goal of providing fewer demonstrations but better performance. At last, by observing the positive relationship between the proportional performance gain and length increase between 2 and 4 max shots scenarios, we conclude that FlexSDR learns the correct time to end the retrieval process and adaptive incorporates the best demonstration for a good marginal performance gain.

\subsection{Ablation Studies}

\label{sec:abla}

\begin{figure}
\centering
\begin{subfigure}{.33\textwidth}
  \centering
  \includegraphics[width=.99\linewidth]{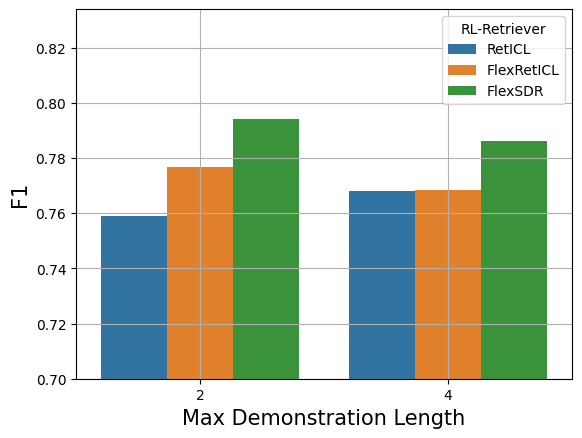}
  \caption{GPT-3.5-turbo}
\end{subfigure}%
\begin{subfigure}{.33\textwidth}
  \centering
  \includegraphics[width=.99\linewidth]{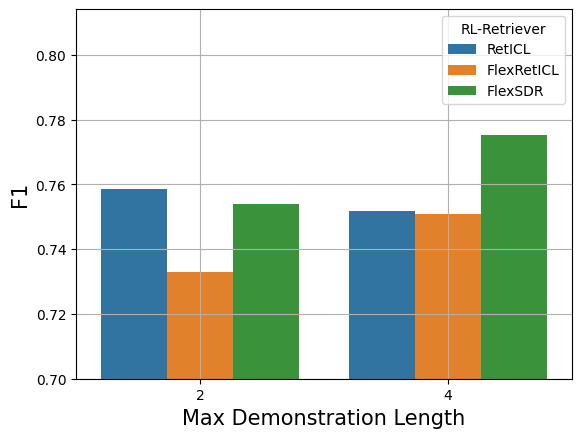}
  \caption{Llama3-8B}
\end{subfigure}
\begin{subfigure}{.33\textwidth}
  \centering
  \includegraphics[width=.99\linewidth]{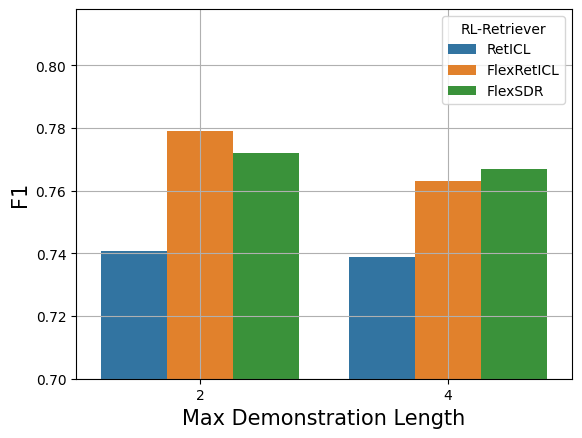}
  \caption{Mistral-7B}
\end{subfigure}
\vspace{-5mm}
\caption{Performance of RetICL, FlexRetICL and FlexSDR with different LLMs.}
\label{fig:ablation}
 \vspace{-5mm}
\end{figure}

To answer RQ3, we ablate the intermediate reward design from FlexSDR and name the new model as FlexRetICR since it is similar to RetICL but can perform the early stop action. We train FlexRetICR with both rewards $r_t$ and $r'_t$ and set weight parameter $\omega = \frac{1}{T}$ since we do not want the accumulated stop bonus reward to become larger than the correctness reward. The return function $R''(\tau)$ for this model is shown as Fig~\ref{fig:return_n} in the Appendix~\ref{sec:return_flexreticr}. For the fair comparison between FlexRetICR and RetICR, we set $\gamma=1$ for FlexRetICR. The performance comparison between the three models is shown as Figure~\ref{fig:ablation}. From the figure, we can observe that FlexRetICR outperforms RetICR in 4 out of 6 cases, which indicates that introducing early stop rewards not only helps to use less demonstrations but also could be beneficial to the final performance. Finally, by comparing FlexSDR with the other two RL-Retrivers, we find that it achieves the best performance in 5 out of 6 scenarios, which proves the effectiveness of the step-wise reward design.

\subsection{Case Studies}

\begin{figure}
\centering
\begin{subfigure}{.33\textwidth}
  \centering
  \includegraphics[width=.99\linewidth]{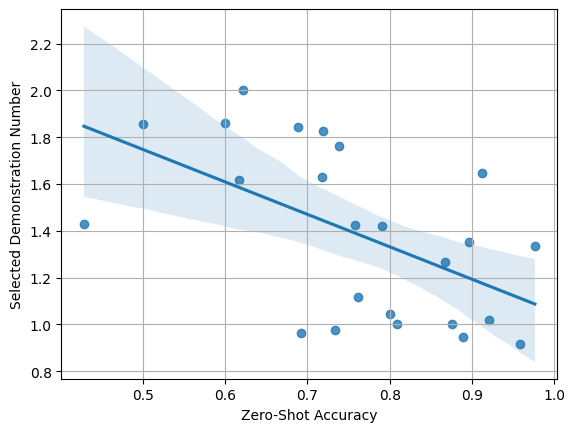}
  \caption{GPT-3.5-turbo}
\end{subfigure}%
\begin{subfigure}{.33\textwidth}
  \centering
  \includegraphics[width=.99\linewidth]{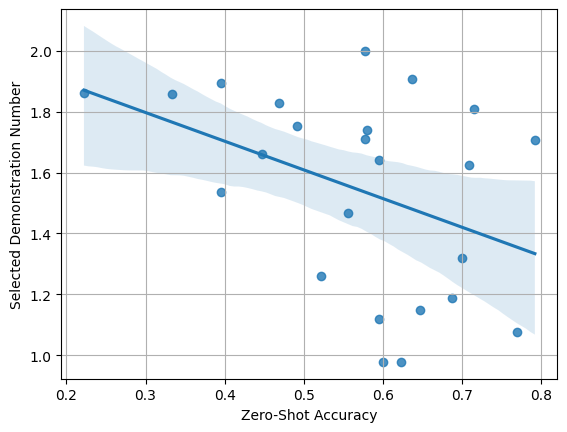}
  \caption{Llama3-8B}
\end{subfigure}
\begin{subfigure}{.33\textwidth}
  \centering
  \includegraphics[width=.99\linewidth]{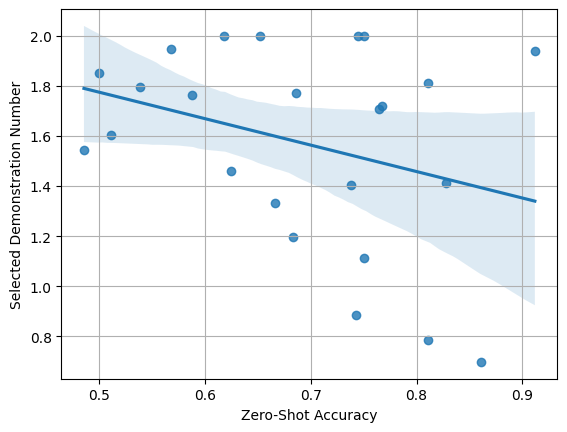}
  \caption{Mistral-7B}
\end{subfigure}
\vspace{-5mm}
\caption{Zero-shot accuracy of different knowledge concepts with corresponding demonstration numbers on different LLMs. Each point in the figure represents a knowledge concept.}
 \vspace{-5mm}
\label{fig:case}
\end{figure}

In this section, we answer RQ4 by presenting the FlexSDR's behavior when facing to knowledge concepts with different zero-shot accuracy. From Figure~\ref{fig:case}, we observe that there is a significant negative relationship between the knowledge-level accuracy at zero-shot performance and the number of demonstrations suggested by FlexSDR. This fact indicates that FlexSDR learns to retrieve fewer demonstrations for knowledge points that already perform well with no demonstration samples. Such a phenomenon provides evidence that FlexSDR learns how to provide an adaptive number of demonstrations to different knowledge concepts.

\section{Conclusion}

In this paper, we present, KnowTS, a LLMs based knowledge-tagging system, which differs from prior machine learning models in its strong performance 
while facing limited or even annotated data for knowledge-tagging tasks. Besides that, we further propose a novel RL-based demonstration retriever, FlexSDR, focusing on dynamically providing flexible lengths of demonstration samples to every question knowledge-matching query. To validate the effectiveness of KnowTS, we experiment with an expertly annotated knowledge concept question dataset, MathKnowCT. The experiment results demonstrate the effectiveness of FlexSDR, which enables KnowTS to achieve the best few-shot learning performance while using fewer demonstrations. At last, through the ablation study and case analyzing results, we demonstrate the effectiveness of each component in FlexSDR.


\bibliographystyle{unsrt}
\bibliography{neurips_2024}

\newpage
\appendix
\section{MathKnowCT Details}
\label{sec:data_detail}
The detailed statistics about MathKnowCT is shown in Tabel~\ref{tab:data_detail}. Overall, there are 2,349 samples covering 24 knowledge concepts of math study for student from Grade 1 to Grade3. The example knowledge definitions and questions are presented in Table~\ref{tab:data_knowledge}. More details can be found from the \url{link}.

\begin{table}
\centering
\caption{Detailed sample statistics for different knowledge concepts in MathKnowCT.}
\label{tab:data_detail}
\resizebox{\textwidth}{!}{
\begin{tabular}{@{}cccc|cccc@{}}
\toprule
\textbf{Knowledge ID} & \textbf{Total Size} & \textbf{Positive Size} & \textbf{Negative Size} & \textbf{Knowledge ID} & \textbf{Total Size} & \textbf{Positive Size} & \textbf{Negative Size} \\ \midrule
 x02030701 & 100 & 25 & 75 & x07020402 & 87 & 29 & 58 \\
x02021101 & 100 & 40 & 60 & x07020502 & 100 & 50 & 50 \\
x06020104 & 100 & 40 & 60 & x20050401 & 100 & 50 & 50 \\
x02061003 & 100 & 16 & 84 & x09020509 & 100 & 50 & 50 \\
 x48040202 & 100 & 29 & 71 & x07020314 & 100 & 30 & 70 \\
x11041602 & 100 & 24 & 76 & x01010201 & 100 & 50 & 50 \\
 x04030501 & 100 & 48 & 52 & x11040205 & 100 & 26 & 74 \\
x04030601 & 100 & 23 & 77 & x11040203 & 100 & 22 & 78 \\
 x07010103 & 100 & 50 & 50 & x11040202 & 100 & 25 & 75 \\
x06030101 & 100 & 44 & 56 & x02040502 & 100 & 44 & 56 \\
 x57130902 & 100 & 35 & 65 & x47060201 & 100 & 17 & 83 \\
x20041003 & 62 & 50 & 12 & x20070401 & 100 & 47 & 53 \\ \bottomrule
\end{tabular}}
\end{table}

\section{Return Function of FlexRetICR}
\label{sec:return_flexreticr}

\begin{figure}[!btph]
\centering
\begin{subfigure}{.45\textwidth}
  \centering
  \includegraphics[width=.99\linewidth]{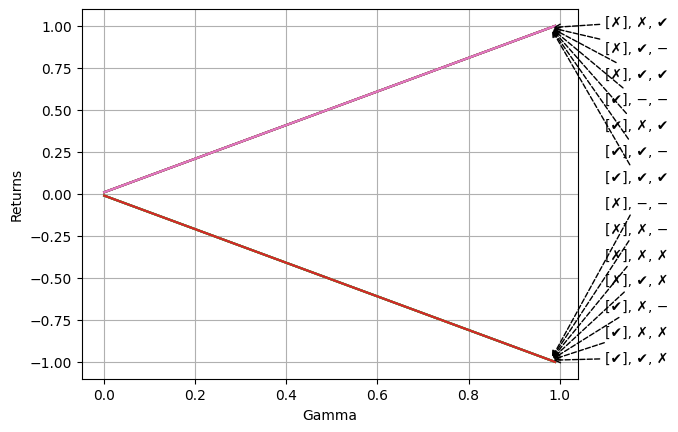}
  \caption{$R(\tau)$ with $\gamma \in (0,1)$}
  \label{fig:return_nob_a}
\end{subfigure}%
\begin{subfigure}{.45\textwidth}
  \centering
  \includegraphics[width=.99\linewidth]{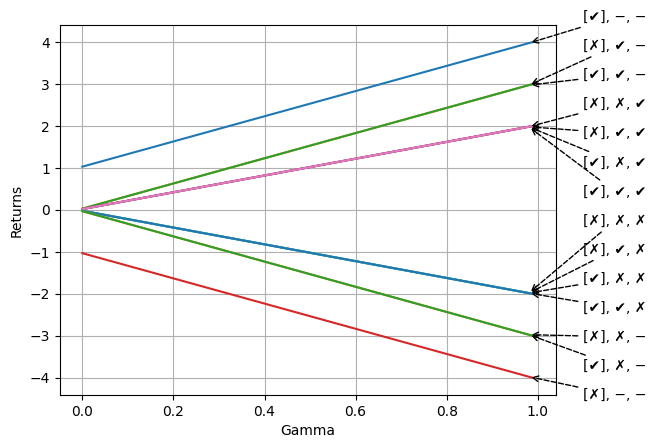}
  \caption{$R''(\tau)$ with $\omega=\frac{1}{2},\ \gamma \in (0,1)$}
  \label{fig:return_b_a}
\end{subfigure}
\caption{Return functions w/o and w/ stop bonus reward for FlexRetICR where $T=2$.}
\label{fig:return_n}
\end{figure}

\section{LLMs Implementation Details}

\label{sec:llm_implement}

To be noticed in this paper, we choose to experiment only with each LLM's instruct-tuned (chat-tuned) version as we observe that the instruct-tuned LLMs can better follow the given annotating instructions and generate the correct format responses. For each framework, we experiment with two-sized versions (small and large) and the prompt text is adjusted based on the preference of each LLM. We run our experiment with 8 * Nvidia A100 80G GPUs.

\begin{table}[]
\caption{Details about LLM implementation in this paper and source file links.}
\resizebox{\textwidth}{!}{
\begin{tabular}{@{}lll@{}}
\toprule
LLM & Model ID & Details \\ \midrule
GPT-Large & gpt-4-turbo-2024-04-09 & \url{https://platform.openai.com/docs/models/gpt-4-turbo-and-gpt-4} \\
GPT-Base & gpt-3.5-turbo-0125 & \url{https://platform.openai.com/docs/models/gpt-3-5-turbo} \\
Llama3-Large & Llama-3-70B-Instruct & \url{https://huggingface.co/meta-llama/Meta-Llama-3-70B-Instruct} \\
Llama3-Base & Llama-3-8B-Instruct & \url{https://huggingface.co/meta-llama/Meta-Llama-3-8B-Instruct} \\
Mixtral-Large & Mixtral-8x7B-Instruct-v0.1 & \url{https://huggingface.co/mistralai/Mixtral-8x7B-Instruct-v0.1} \\
Mixtral-Base & Mistral-7B-Instruct-v0.2 & \url{https://huggingface.co/mistralai/Mistral-7B-Instruct-v0.}2 \\
Qwen1.5-Large & Qwen1.5-72B-Chat & \url{https://huggingface.co/Qwen/Qwen1.5-72B-Chat} \\
Qwen1.5-Base & Qwen1.5-7B-Chat & \url{https://huggingface.co/Qwen/Qwen1.5-7B-Chat} \\ \bottomrule
\end{tabular}}
\end{table}

\begin{table}[]
\caption{Example knowledge definitions}
\label{tab:data_knowledge}
\begin{tabular}{|p{0.2\textwidth}|p{0.8\textwidth}|}
\toprule
Knowledge ID & Knowledge Definition\\ \midrule
x01010201 & Learn the definitions of the following types of numbers, including   integers, odd numbers, even numbers, fractions, decimals, positive numbers,   negative numbers, and natural numbers. Common related question types include   the following: (1) Select a number of a specified type from a given set of   numbers; (2) Determine whether a number is within the defined range; (3)   Determine whether a proposition about the classification of numbers is true   or false. \\ \midrule
x02021101 & Learn the relationship between natural numbers within 5. Common related   question types include the following: (1) Sorting numbers by size; (2)   Comparing numbers. Also note: the question stem does not contain addition,   subtraction, multiplication, or division formulas. \\ \midrule
x02040502 & Learn the composition of two-digit numbers less than or equal to 100 (how   many tens and how many ones). Common related question types include the   following: (1) Convert a two-digit number into a combination of tens and   ones; (2) Fill in the corresponding two-digit number based on the combination   of tens and ones. \\ \midrule
x02061003 & Learn to use 3 or 4 digits to form a three-digit or four-digit number,   and judge the size relationship between the digits. Related question types   are limited to the following: (1) Use 3 digits to form a three-digit number   smaller than a certain number. Find the total number of such three-digit   numbers, the largest number, and the smallest number. Each digit can only be   used once in the combination process. (2) Knowing that the sum of the digits   in each digit of a four-digit number is a certain number, find the largest   number and the smallest number of this four-digit number. \\ \midrule
x04030501 & Learn to calculate the reciprocal of a number. Common related question   types include the following: (1) Calculate the reciprocal of one or more   given numbers; (2) Given an equation where the product of a number and a   blank is 1, find the value of the number that can be filled in the blank. \\ \midrule
x48040202 & Learn how to estimate the total purchase price of three items in a   shopping scenario. Common related question types include the following: (1)   Given the prices of three items (each item can be a three-digit or two-digit   price), but at least one of the items has a three-digit price, calculate the   approximate total purchase price of the three items; (2) Calculate both the   approximate and exact total purchase price of the three items; \\ \midrule
x57130902 & Learn to solve feasible combinations by enumeration. Common related   question types include the following: (1) Given a numerical value of a total   quantity demanded (e.g., total quantity of goods transported, total price),   and the numerical value that each option can provide (e.g., the loading   capacity of trucks of different sizes, coins of different denominations),   solve the option combination that just meets the total quantity demanded.   Also note that the numbers in the question stem are all integers, and the   numerical value of each option in the combination cannot be wasted (e.g.,   each truck must be fully loaded, and no change is given for the currency). In   the problem-solving process, no more than 15 feasible combinations should be enumerated. \\ \bottomrule
\end{tabular}
\end{table}

\end{document}